\documentclass[conference]{IEEEtran}
\usepackage{cite}  
\usepackage{amsmath,amssymb,amsfonts}
\usepackage{algorithm}
\usepackage{algorithmicx}
\usepackage{algpseudocode}
\usepackage{algpseudocodex}
\usepackage{graphicx}
\usepackage{textcomp}
\usepackage{xcolor}
\def\BibTeX{{\rm B\kern-.05em{\sc i\kern-.025em b}\kern-.08em
    T\kern-.1667em\lower.7ex\hbox{E}\kern-.125emX}}
    
\usepackage{todonotes}

\DeclareMathOperator*{\argmax}{arg\,max} 
    
\begin{document}

\title{Solving the Kidney-Exchange Problem via Graph Neural Networks with No Supervision}

\newcommand{\temqver}[1]{\colorbox{red}{\textcolor{yellow}{\textbf{#1}}}}

\author{
\IEEEauthorblockN{Pedro Foletto Pimenta}
\IEEEauthorblockA{\textit{Institute of Informatics} \\
\textit{UFRGS Federal University} \\
Porto Alegre, RS, Brazil \\
pfpimenta@inf.ufrgs.br} 
\and
\IEEEauthorblockN{Pedro H. C. Avelar}
\IEEEauthorblockA{\textit{King's College London} \\
London, UK \\
pedro\_henrique.da\_costa\_avelar@kcl.ac.uk \\
\textit{A*STAR} \\
Singapore
}
\and
\IEEEauthorblockN{Luis C. Lamb}
\IEEEauthorblockA{\textit{Institute of Informatics} \\
\textit{UFRGS Federal University} \\
Porto Alegre, RS, Brazil \\
lamb@inf.ufrgs.br}
}

\maketitle

\begin{abstract}
This paper introduces a new learning-based approach for approximately solving the Kidney-Exchange Problem (KEP), an NP-hard problem on graphs.
The problem consists of, given a pool of kidney donors and patients waiting for kidney donations, optimally selecting a set of donations to optimize the quantity and quality of transplants performed while respecting a set of constraints about the arrangement of these donations.
The proposed technique consists of two main steps:
the first is a Graph Neural Network (GNN) trained without supervision;
the second is a deterministic non-learned search heuristic that uses the output of the GNN to find a valid solution.
To allow for comparisons, we also implemented and tested an exact solution method using integer programming, two greedy search heuristics without the machine learning module, and the GNN alone without a heuristic.
We analyze and compare the methods and conclude that the learning-based two-stage approach is the best solution quality, outputting approximate solutions on average 1.1 times more valuable than the ones from the deterministic heuristic alone.
\end{abstract}


\section{Introduction}
\label{introduction}

This study addresses machine learning approaches for the approximate solving of the Kidney Exchange Problem (KEP), an NP-Hard problem on graphs \cite{kidney_exchange_2004,KEP_2015}.
This problem consists of, given a pool of kidney donors and patients waiting for kidney donations, optimally selecting a set of donations to optimize the quantity and quality of transplants performed while still respecting a set of constraints about the arrangement of these donations.


Thus, this work's main objective is to answer the following question: \textbf{Can the Kidney Exchange problem be better approximately solved with the help of machine learning?}
If positive, we want to evaluate the feasibility of utilizing such an approach in terms of the quality of the solutions it provides.
Further, we are also interested in assessing how viable would such a method be in terms of computational time.
Additionally, one hopes that, by answering these questions, we may also better understand the limitations of the employed machine learning methods for this problem and the potential future research directions for solving the KEP and other optimization problems in graphs.

\subsection{On Kidney Exchanges}
Kidney disease affects millions of people worldwide, and the two available treatment options for end-stage kidney disease are dialysis and kidney transplantation \cite{improved_kep_instance_generation_2022}.
Transplantation is the preferred treatment for the most severe forms of kidney disease \cite{kidney_exchange_2004} because it is cheaper and offers a better quality of life and better life expectancy \cite{economic_assessment_kidney_transplant_2018}.
The source of the kidney can be either a cadaver or a live donor, as the human body has two kidneys, and often only one suffices.

The compatibility of a transplant between a donor and a recipient is determined by a number of different factors, such as the blood-group compatibility, tissue-type compatibility, the ages and general health of the donor and the recipient, the size of the donor's kidney, and many others \cite{improved_kep_instance_generation_2022}.
The lower the compatibility between a donor and a patient, the lower the chance of kidney transplant success between them.

In the last decades, there have started to be \textit{paired kidney exchanges}, which are cycles involving donor-patient pairs such that each donor cannot give a kidney to their intended recipient because of some incompatibility. However, each patient can receive a kidney from a donor from another pair \cite{kidney_exchange_2004}.
These cycles were first performed with only two donor-patient pair nodes, but later longer cycles of kidney exchanges were performed.
Another possibility of an exchange scheme is to create exchange chains that begin with a donation of an altruistic or cadaveric kidney donor, followed by chained donations of patient-donor pairs, and then finish with a donation either to a patient with or without an associated donor.
In this study, the donation chains are also referred to as \textit{paths}, a term often used for describing sequences of connected nodes in graph problems.
To find the best possible allocation, i.e., the optimal solution to the problem, considering a set of donors, patients, and patient-donor pairs, a mix of both cycles and chains can be selected as long as the cycles and paths do not intersect with each other.

These cycles and chains could have unlimited size.
In real life, however, there is a practical limit to the size of the paired kidney exchange cycles and chains:
The kidney donation surgeries in a chain often must be done simultaneously to ensure every patient receives a kidney before her associated donor donates her kidney.
However, organizing many simultaneous surgeries is logistically complex and sometimes impractical or even impossible.
Even if they do not have to be done simultaneously, it is generally required at least that every patient-donor pair receive a kidney before they give a kidney.
Furthermore, numerous other logistical difficulties arise when dealing with longer cycles and chains, which makes it highly desirable or sometimes even necessary that these donation cycles and chains have limited size.
The longest kidney transplant chain successfully performed had a size of 35 and happened between 6 January and 17 June 2015 in the USA \cite{longest_KEP_chain}, although, in most situations, the maximum reasonable size is considerably smaller.

\subsection{The Kidney Exchange Problem}
\label{kep_definition}

The Kidney Exchange Problem (KEP) was first mathematically formalized by Roth et al. in \cite{kidney_exchange_2004}, then slightly updated in various ways in subsequent works.
A summary of the variations found in the literature and models and techniques currently employed to solve them can be found at \cite{biro_2021}.

In the formalization used in the study, each instance of the KEP is represented by a directed weighted graph $G = \{V, E\}$.
Each patient, donor, and patient-donor pair is mapped to a graph node; they will be referred to as patient (P) nodes, non-directed (or altruistic) donor (NDD) nodes, and patient-donor pair (PDP) nodes.
The set of nodes $V$ is thus accordingly partitioned into sets $P$, $NDD$ and $PDP$.
The graph's edges represent donation compatibility: an edge from node A to node B represents that a kidney donation in this direction is possible; the edge weight encodes the donor's compatibility with the recipient.

Solving a KEP instance means optimally selecting a set of cycles and chains to optimize the transplants performed.
This optimally includes maximizing the quantity and quality of the transplants, which is encoded in the edge weights.
The solution must also respect a set of constraints.
Each node may participate at most in one transplant as a donor and  another as a receiver.
Also, PDP nodes can only donate a kidney if they receive one, although they can receive it without donating.
Nodes of type P can only receive donations, and NDD nodes can only donate.
The problem is then described as follows: 
"\textit{Given a list of kidney needing patients, kidney donors, and patient-donor pairs, and a compatibility index between each possible donor and receiver, what is the best possible selection of donations that can be performed so that the total quantity of transplants, weighted by the compatibility indexes, is maximized, while still respecting a given size limit to the kidney exchange cycles and chains in the solution?}"

Our KEP formalization is represented in the set of equations below, inspired on the so-called Recursive Algorithm formulation described by \cite{KEP_2015}.
We use a binary variable $y_e$ for each edge $e \in E$, that indicates if the edge is part of the solution or not, as well as auxiliary variables flow in $f_{v}^{i}$ and flow out $f_{v}^{o}$ for each node $v \in V$, which represents the node's number of incoming and outcoming edges contained in the solution, and are defined at Equations \ref{eq:flow-in} and \ref{eq:flow-out}.
$\mathcal{C}$ represents the set of existing cycles and paths in graph, where $C \in \mathcal{C}$ is a collection of edges, i.e. $C \subset E$.
$\mathcal{C}_{k}$ is a subset of $\mathcal{C}$ (i.e. $\mathcal{C}_{k} \subset \mathcal{C}$) containing the cycles and paths that use $k$ or fewer edges.

\begin{equation}
\begin{aligned}
    \max & \sum_{e \in E}{w_{e}y_{e}} \label{eq:opt-objective}
\end{aligned}
\end{equation}
\begin{align}
    \operatorname{s.t.} ~~ \sum_{e \in \mathcal{N}_{in}(v)}{y_{e}} = f_{v}^{i} ~~~~ v &\in V \label{eq:opt-constraint-flow-in} \\
    \sum_{e \in \mathcal{N}_{in}(v)}{y_{e}} = f_{v}^{o} ~~~~ v &\in V  \label{eq:opt-constraint-flow-out} \\
    f_{v}^{o} \leq f_{v}^{i} \leq 1 ~~~~ v &\in {PDP} \label{eq:opt-constraint-flow-pdp} \\
    f_{v}^{o} \leq 1 ~~~~ v &\in NDD \label{eq:opt-constraint-flow-n} \\
    f_{v}^{i} \leq 1 ~~~~ v &\in P \label{eq:opt-constraint-flow-p} \\
    \sum_{e \in C}{y_{e}} \leq |C|-1 ~~~~ C &\in \mathcal{C} \setminus \mathcal{C}_{k} \label{eq:opt-constraint-cycle-size} \\ 
    y_{e} \in \{0,1\} ~~~~ e &\in E \label{eq:opt-constraint-y-domain}
\end{align}

\begin{equation}
    f_{v}^{i} = \sum_{e \in \mathcal{N}_{in}(v)} y_e ~~~~ v \in V \label{eq:flow-in}
\end{equation}
\begin{equation}
    f_{v}^{o} = \sum_{e \in \mathcal{N}_{out}(v)} y_e ~~~~ v \in V \label{eq:flow-out}
\end{equation}

\[\mathcal{N}_{in}(v) = \{e \forall e \in E, e=(v',v)\}\]
\[\mathcal{N}_{out}(v) = \{e \forall e \in E, e=(v,v')\}\]

The constraints ensure the result is a valid solution for KEP:
the first two (Eq.~\ref{eq:opt-constraint-flow-in} and Eq.~\ref{eq:opt-constraint-flow-out}) are necessary for the use of the flow in and flow out variables,
the third one (Eq.~\ref{eq:opt-constraint-flow-pdp}) controls the flow in and flow out of the PDP nodes,
the fourth and fifth ones (Eq.~\ref{eq:opt-constraint-flow-n} and Eq.~\ref{eq:opt-constraint-flow-p}) do the same but for NDD nodes and P nodes, respectively,
the sixth one (Eq.~\ref{eq:opt-constraint-cycle-size}) prohibits cycles or paths with length longer than a given limit $k$,
and the seventh one (Eq.~\ref{eq:opt-constraint-y-domain}) defines the domain of the \textbf{y} variable.
The objective (defined at Expression \ref{eq:opt-objective}) is to maximize the number of edges in the solution $y$, weighted by the associated edge weights $w$, while still respecting the KEP constraints (Equations \ref{eq:opt-constraint-flow-pdp}, \ref{eq:opt-constraint-flow-n}, \ref{eq:opt-constraint-flow-p}, \ref{eq:opt-constraint-cycle-size}, and \ref{eq:opt-constraint-y-domain}).

It has been proven that this problem is NP-Hard \cite{abraham_2007}, although it can become polynomial-time solvable if some of the constraints are relaxed, such as limiting the exchange cycles and chains length to 2, or removing the length restriction entirely.


\section{Machine Learning Methods for Optimization Problems in Graphs}
\label{ml_for_optimization_in_graphs}

In the last few years, many machine learning-based approaches that effectively solve several different optimization problems in graphs have been proposed, although none of them designed for solving KEP.
Graph optimization problems already solved with the help of machine learning include the Set Covering Problem \cite{csc_tsp_ml_2020}, Graph Colouring \cite{graph_coloring_gnn_2019, graph_coloring_gnn_2020}, Minimum Vertex Cover \cite{gnn_minimum_vertex_cover_2019, gnn_np_hard_alpha_go_2019}, Maximum Cut \cite{learning_comb_opt_graphs_2017}, Graph Partitioning \cite{graph_partitioning_ml_2019}, Maximum Independent Set \cite{gnn_maximum_independent_set_2018}, Maximum Common Subgraph \cite{gnn_maximum_common_subgraph_2021}, and the Travelling Salesperson Problem (also called the travelling salesman problem or TSP), one of the most famous NP-hard problems, often used to represent the class, and some variants of it \cite{gnn-beam_search_2019, learning_tsp_2022, prates_2018, pointer_networks_2015, learning_improvement_heuristics_tsp_2019, attention_learn_tsp_2018}.

In 2015, the authors of \cite{pointer_networks_2015} presented a new type of neural network called Pointer Networks, designed to learn how to reorder the elements of an input sequence; they validated the method by using it to solve 3 problems, including the TSP.
In 2016, researchers presented in \cite{bengio_2016} a framework for combinatorial optimization problems using neural networks and reinforcement learning; the work focused on the TSP, which is a graph problem, but the approach was designed to work with any combinatorial optimization problem.
In 2017, the authors of \cite{learning_comb_opt_graphs_2017} proposed the utilization of a combination of graph representation learning and reinforcement learning to solve graph optimization problems; they showed that their proposed approach effectively learns to solve at least three of those problems: Minimum Vertex Cover, Maximum Cut, and TSP.
In 2018, the decision variant of the TSP, called Decision Traveling Salesman Problem (DTSP), which is to decide if a given TSP instance admits a Hamiltonian route with a cost no greater than a given threshold C, and is also NP-Hard, was solved in \cite{prates_2018} with a GNN.

In 2019, the authors of \cite{gnn-beam_search_2019} tried to solve the TSP using a two stage technique that is very similar to the one presented in this study (described at section \ref{two_stage_method} and illustrated at Figure \ref{images/2_stage_method.drawio.png}): firstly, a GNN processes the input graph and create scores for each edge of the graph; then, a non-learned search heuristic, which in this case was beam search, uses these scores to construct a solution.
There have been other approaches that use similar techniques: in \cite{graph_learning_comb_opt_survey_2020} the authors review graph learning methods for solving CO problems, with a focus on two-stage techniques, where the first is based on graph representation learning, which embeds the input graph into low-dimension vectors, and the second uses the embeddings learned in the first stage; \cite{gnn_nsc_survey_2020} surveys the use of GNNs as a model of neural-symbolic computing and their applications, which includes combinatorial optimization problems; 
\cite{learning_tsp_2022} unifies and refines several of such two stage techniques for neural combinatorial optimization, and test it on the TSP.


\section{Dataset} \label{dataset}

Deep learning methods such as GNNs require lots of data; usually, many thousands of instances, or even millions, depending on the problem and the size of the model, are necessary for the models to converge to a decent behaviour.
With insufficient data, the model is very prone to overfitting, or sometimes may not even learn anything at all.
Medical data, however, is very scarse, and rarely available at this quantity.
This happens for two main reasons.
Firstly, a major problem with healthcare data is its sensitivity: as a lot of it is confidential information about the patients, it is usually highly protected and as a rule cannot be used without special consent, be it from the patients or at least from the health institute that owns the data.
Furthermore, there are a limited number of medical cases of each given situation registered; although it would be useful to have a KEP dataset with millions of instances, this situation has not happened that many times, and not necessarily all of them have been registered digitally.

Due to the scarcity of the data, and considering that to train a machine learning model it usually takes at least tens of thousands of examples, the datasets were generated artificially.
Three separate datasets were generated:
the \textit{train dataset}, with 10 thousand instances for the training of the model;
the \textit{validation dataset} with 100 instances, used for validation step during the training;
and the \textit{test dataset}, with 10 thousand instances, used to evaluate the performance of each one of the employed methods.
This number of instances for the validation dataset was chosen because it was sufficiently big so that still kept roughly the same properties as the train and test datasets, but small enough that the validation does not slow down the training too much.



To generate each instance, first 300 separated nodes are created, and then 5500 edges are added sequentially linking random nodes, while still guaranteeing that no two edges connect the same two nodes in the same direction.
The weight value of each edge is sampled from an uniform distribution of values between 0 and 1.
To choose the number of nodes and edges of the KEP generated instances, the instances used for benchmarking in \cite{KEP_2015} were used as reference.
Considering the 25 instances presented in the Table S3, which they call "difficult" real-data instances, the average number of nodes on a KEP instance is 265.84, and the average number of edges is 5695.92.
Thus, the values for the number of nodes and number of edges chosen were 300 and 5500, respectively.
The proportion of the types of nodes was also chosen to be similar to the instances: roughly 90\% of the nodes are PDP nodes, 5\% are NDD nodes, and 5\% are P nodes.

To keep track of the instances, an unique ID was assigned to each.
For this, the Weisfeiler Lehman graph hash \cite{weisfeiler_lehman_graph_kernels} was used, which guarantees that non-isomorphic graphs will get different hashes;
it also considers node and edge features in its computation, thus differentiating even between graphs which have the same structure but different edge weight values.

\section{Methods}
We classified the methods for solving the KEP in 3 categories: integer programming methods, non-learnable heuristic methods, and learnable heuristic methods.
All of them use the same input information, which is a KEP instance, and return the solution in the same format, which is a binary label for each edge, indicating if it is in the solution or not.
All of them are evaluated on the test dataset, with the exception of the integer programming method, as later explained in Section \ref{experiments}, but only the learned heuristics use the training dataset, during their training phase.

\subsection{Integer Programming}
\label{integer_programming}

To obtain the analytical solution, i.e. the optimal solution, the formulation presented in Subsection \ref{kep_definition} was implemented using the PyCSP3 Python library \cite{pycsp3}.
There are other integer programming formulations for KEP, including two presented in \cite{KEP_2015}, as well as others in \cite{roth_2007}, \cite{abraham_2007} and \cite{kep_ip_models_2013}.
This formulation was chosen because it is the most straightforward one.

\subsection{Non-Learnable Heuristics}
\label{non-learnable_heuristics}

To evaluate the implemented methods that use machine learning, we decided to compare them to non-learnable heuristics, i.e. heuristic methods that do not use learning techniques.
This section aims to describe these non-learnable heuristic methods.
To the best of our knowledge, however, there are no canonical heuristics for the KEP.
For this reason, we implemented two search heuristics, which are described below.

\subsubsection{Greedy Paths}
\label{greedy_paths}
This algorithm greedily selects paths that start on NDD nodes and goes through PDP or P nodes one by one until there is no more nodes to be selected, or until a P node is reached.
It starts by selecting the edge with the highest weight considering only the subset of edges that have an NDD node as source.
Then, considering only the edges that come from the previous node, it follows by selecting always the next edge with the highest weight, until there are no more available edges left that would continue the path.
After a path has been added to the solution, the edges connected to nodes of this path are masked, and Greedy Paths repeats the process until no more NDD nodes with valid outgoing edges are available.
This algorithm is described at Algorithm \ref{alg:greedy-paths}.


\begin{algorithm}
\caption{Greedy-Paths}
\label{alg:greedy-paths}
\begin{algorithmic}
\Procedure{Greedy-Paths}{$G=(N,E),k$}
    \State $\operatorname{paths} \gets []$
    \While{$|\operatorname{GP}(G,k)| > 0$}
        \State $\operatorname{path} \gets \operatorname{GP}(G,k)$
        \State $\operatorname{paths} \gets \operatorname{paths} \oplus [\operatorname{path}]$
        \State $G \gets (N \setminus \{n ~\forall~ n \in \operatorname{paths}\}, E \setminus \{e ~\forall~ e \in E, src(e)=n \lor tgt(e)=n\} )$
    \EndWhile
    \State \textbf{return} $\operatorname{paths}$
\EndProcedure
\Procedure{GP}{$G=(N,E),k$} \Comment{Gets one Greedy Path}
    \State $E_{NDD} \gets \{ e ~\forall~ e \in E, src(e) \in NDD \}$
    \If{$|E_{NDD}|=0$}
        \State \textbf{return} $[]$
    \EndIf
    \State $e_{c} \gets \argmax_{e \in E_{NDD}\}} w_{e}$
    \State $\operatorname{path} \gets [src(e_{c})]$
    \While{$|\operatorname{OE}(tgt(e_{c}))|>0 \land |\operatorname{path}|<(k + 1)$}
        \State $\operatorname{path} \gets \operatorname{path} \oplus [tgt(e_{c})]$
        \State $e_{c} \gets \argmax_{
            e \in
                \operatorname{OE}(tgt(e_{c}))
                \setminus \bigcup_{
                    n \in \operatorname{path}
                } \operatorname{IE}(n)
        } w_{e}$
    \EndWhile
    \State \textbf{return} $\operatorname{path}$
\EndProcedure
\Procedure{oe}{$G=(N,E),n$} \Comment{Outgoing Edges}
    \State \textbf{return} $\{e ~\forall~ e \in E, src(e)=n\}$
\EndProcedure
\Procedure{ie}{$G=(N,E),n$} \Comment{Incoming Edges}
    \State \textbf{return} $\{e ~\forall~ e \in E, tgt(e)=n\}$
\EndProcedure
\end{algorithmic}
\end{algorithm}

\subsubsection{Greedy Cycles}
\label{greedy_cycles}

This algorithm greedly selects cycles of PDP nodes.
It starts by selecting the edge with the highest weight considering only the subset of edges that have a PDP node as source and a PDP node as destination.
Then, PDP nodes are greedly added to the solution in the same way as done by the Greedy Paths method until the cycle ends, or until it arrives at a node already added in the cycle, in which case the cycle is closed and the nodes before the current node are removed from the cycle.
After a cycle is added to the solution, the Greedy Cycles algorithm applies a mask on the edges connected to nodes of this cycle, and then repeats the process until no more PDP nodes with valid outcoming edges are available.
This algorithm is described at Algorithm \ref{alg:greedy-cycles}.

\begin{algorithm}
\caption{Greedy-Cycles}
\label{alg:greedy-cycles}
\begin{algorithmic}
\Procedure{Greedy-Cycles}{$G=(N,E),k$}
    \State $\operatorname{cycles} \gets []$
    \While{$|\operatorname{GC}(G,k)| > 0$}
        \State $\operatorname{cycle} \gets \operatorname{GP}(G,k)$
        \State $\operatorname{cycles} \gets \operatorname{cycles} \oplus [\operatorname{cycle}]$
        \State $G \gets (N \setminus \{n ~\forall~ n \in \operatorname{cycles}\}, E \setminus \{e ~\forall~ e \in E, src(e)=n \lor tgt(e)=n\} )$
    \EndWhile
    \State \textbf{return} $\operatorname{cycles}$
\EndProcedure
\Procedure{GP}{$G=(N,E),k$} \Comment{Gets one Greedy Cycle}
    \State $E_{PDP} \gets \{ e ~\forall~ e \in E, src(e) \in PDP, dst(e) \in PDP \}$
    \If{$|E_{PDP}|=0$}
        \State \textbf{return} $[]$
    \EndIf
    \State $e_{c} \gets \argmax_{e \in E_{PDP}\}} w_{e}$
    \State $\operatorname{cycle} \gets [src(e_{c})]$
    \While{$tgt(e_{c}) \notin \operatorname{cycle}$}
        \If{$|\operatorname{OE}(tgt(e_{c}))| \le 0 \lor |\operatorname{cycle}| \ge k$}
            \State \textbf{return} $[]$
            \Comment{Unable to close cycle (dead end)}
        \EndIf
        \State $\operatorname{cycle} \gets \operatorname{cycle} \oplus [tgt(e_{c})]$
        \State $e_{c} \gets \argmax_{
            e \in
                \operatorname{OE}(tgt(e_{c}))
                \setminus \bigcup_{
                    n \in \operatorname{cycle}
                } \operatorname{IE}(n)
        } w_{e}$
    \EndWhile
    \State \textbf{return} $\operatorname{cycle}$
\EndProcedure
\Procedure{oe}{$G=(N,E),n$} \Comment{Outgoing Edges}
    \State \textbf{return} $\{e ~\forall~ e \in E, src(e)=n\}$
\EndProcedure
\Procedure{ie}{$G=(N,E),n$} \Comment{Incoming Edges}
    \State \textbf{return} $\{e ~\forall~ e \in E, tgt(e)=n\}$
\EndProcedure
\end{algorithmic}
\end{algorithm}

\subsection{Learnable Heuristics}
\label{learnable_heuristics}


As KEP instances are graphs, using GNNs to extract more detailed and abstract information can help in constructing an approximate solution. 
Therefore, GNN models were chosen as the main learning module for the machine learning methods.

\subsubsection*{GNN Architecture}
\label{gnn_architecture}

The architecture of the GNN used in this work is the following:
firstly, there is a message passing phase, where the original node features are passed through a PNA layer \cite{pna_2020}, and then through two consecutive GATv2 layers \cite{gat_v2_2021};
after each message passing layer, a ReLU activation function is applied, followed by a dropout regularization;
next, the node features are passed through a fully connected feed forward neural network, followed by another ReLU activation function;
then, the edge features are constructed by concatenating the original input edge features with the node features of the origin and destination nodes associated to each edge;
these edge features are then passed through a fully connected feed forward neural network, which outputs a score for each edge;
at this point, a skip connection adds the original edge weights to the edge scores;
finally, a node-wise softmax operation, which is described below, is applied so as to normalize these scores in relation to the scores of other edges that share the same source node.

In order to make information flow not only in the original direction of the original edges, each message passing layer is accompanied by an associated layer, which we call \textit{counter edge layer}, that is exactly similar, but with different learned weights, and with the difference that the information is propagated in the opposite direction, i.e. flowing from the destination node to the origin node.
Each time message layers are executed, the output of the original and of the counter edge layers is concatenated before being passed to the next layers.

\subsubsection*{KEP Unsupervised Loss}
\label{kep_unsupervised_loss}

This loss function was designed to capture, without the need for the exact solution as a label or any other supervision, the essence of what we are trying to maximize: the sum of weights of edges that are in the predicted solution.
It is defined as the log of the sum of weights of all edges of the input instance over the sum of weights of edges that are in the predicted solution, weighted by the scores predicted by the GNN.
This loss function is presented in Formula \ref{loss_formula}, where $w$ represents the vector of edge weights, $pred$ represents the vector of predicted classes (which indicates if each edge is contained in the solution or not), and $s$ represents the vector of scores attributed by the GNN for each edge.


\begin{eqnarray}
    KEP\_ Loss \left ( w, pred, s \right )  = \log \frac{\sum_{e \in E}^{}w_e}{\sum_{e \in E}^{} w_e pred_e s_e}
    \label{loss_formula}
\end{eqnarray}

\subsubsection*{Loss Constraint Regularization}
\label{kep_loss_constraint_regularization}


In order to integrate information about the KEP flow constraints (Eq. \ref{eq:opt-constraint-flow-pdp}, Eq. \ref{eq:opt-constraint-flow-n} and Eq. \ref{eq:opt-constraint-flow-p}) into the learning of the model, a loss regularization function was developed.
It aims to model the restriction that each node must have at maximum one single outcoming edge and one incoming edge that are part of the solution.
It is defined as the log of the division between the total quantity of edges in the solution and the number of unique nodes that appear in the solution as a origin/destination node.
This makes it so that the regularization term value is proportional to the number of invalid edges in the solution, i.e. the total quantity in the graph of extra edges for each source/destination node.
This function can then be added to the unsupervised loss (described in the subsection above) by summing their output values, weighted by coefficients, which become new hyper-parameters of the training.

\subsubsection*{Node-wise Softmax}
\label{node_wise_softmax}

The node-wise softmax operation is the application of an independent softmax operation for each group of edges that share the same source node, as one can see in Equation~\ref{eq:nwsoftmax}, where $s_{e_{(n_{i},n_{j})}}$ represents the edge score of the edge connecting node \textit{i} to node \textit{j}.
In this way, for each node we will have a probability for each outgoing edge; in KEP, these values may represent a probability distribution for the donation options of the donor for each source node.
Although in this work the operation was used grouping edges by source node, with a simple change of a parameter it can group edges by groups of common destination node as well.

\begin{equation}\label{eq:nwsoftmax}
    \operatorname{node-wise-softmax}(s_{e_{(n_{i},n_{j})}}) = \frac{e^{-s_{e_{(n_{i},n_{j})}}}}{\sum_{n_{k} \in \mathcal{N}(n_{i})}{e^{-s_{e_{(n_{i},n_{k})}}}}}
\end{equation}



To the best of our knowledge, this is the first time this operation is proposed in the literature.
It is potentially useful for any edge classification task on graphs, specially when the problem involves constraints in which only one edge may be chosen per node, be it destination or origin node.
These constraints are very common in optimization problems in graphs; this is the case for KEP, for example: each donor or patient-donor pair may donate at most one kidney, and each patient or patient-donor pair may receive at most one kidney.

\subsubsection{Unconstrained GNN Model}
\label{unconstrained_gnn_unsupervised}


This method, referred to from now on as \textit{Unsupervised GNN}, consists of a GNN model which receives a KEP instance and outputs, for each edge of the input instance, a score and a binary prediction, which indicates if the edge is part of the predicted solution or not.
The binary prediction is made independently for each edge, and consists of a simple decision threshold.
This GNN model is trained using the unsupervised loss described at subsection \ref{kep_unsupervised_loss} with the loss regularization term described at subsection \ref{kep_loss_constraint_regularization}.
Although there is no guarantee that the solutions given by this method will be valid, the loss regularization term is used with the goal of inducing it to respect the problem constraints.

\subsubsection{Two Stage Method}
\label{two_stage_method}

\begin{figure}
    \centerline{\includegraphics[width=88mm]{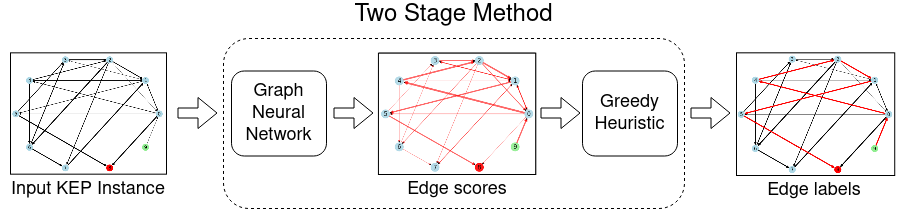}}
    \caption{Diagram representing an overview of the \textit{two stage method}. The GNN takes the input KEP instance and computes a score for each edge of the graph; then, a greedy heuristic such as \textit{GreedyCycles} or \textit{GreedyPaths} uses these edge scores instead of the original edge weights to build an approximate solution, which is a binary label for each edge (\textit{edge labels}), indicating if the edge is part of the approximate solution predicted or not.
    Source: Author.}
    \label{images/2_stage_method.drawio.png}
\end{figure}

Inspired by the approach used in \cite{gnn-beam_search_2019} and \cite{learning_tsp_2022},  this method follows a two steps structure:
firstly, the learnable step, which is a GNN, takes the KEP graph instance as an input and outputs a score for each edge;
then, the non-learnable step, which is one of the heuristic methods described above in the \ref{non-learnable_heuristics} section is executed, but using the scores given by the GNN instead of the edge weights.
This process is illustrated in the diagram on Figure \ref{images/2_stage_method.drawio.png}.
The intuition behind this idea is that the GNN model will learn to encode in the edge score contextual information that will change the decisions of the search heuristic so as to maximize the total score of the final output solution.

This GNN model is trained without supervision using loss described at subsection \ref{kep_unsupervised_loss}.
There is no need to use the loss regularization term described at subsection \ref{kep_loss_constraint_regularization} because the second step of the method ensures that the output will be a valid solution.

Two versions of the two stage method were implemented for the experiments.
Both use the GNN described in subsection \ref{gnn_architecture} for the first stage, but their second stage consist of different search heuristics.
One uses the Greedy Paths search heuristic described at subsection \ref{greedy_paths} and is referred to later on as \textit{GNN+GreedyPaths}.
The other uses the Greedy Cycles search heuristic described at subsection \ref{greedy_cycles} and is referred to later on as \textit{GNN+GreedyCycles}.

\section{Experiments}
\label{experiments}

The main goal of the experiments was to assess the machine learning methods, and compare them to the deterministic heuristics and to the exact solution, both in terms of the quality of the solutions as well as of their operational performance, i.e. the time it takes for a solution to be calculated.
The integer programming method, however, could not be measured in the same dataset because it took to long to run the solver.



The objective of the Kidney Exchange Problem is to maximize the number of donations weighted by their compatibility index, i.e. their associated edge weights in the graph.
Therefore, this was the main metric used to quantify the quality of each solution and to compare the different methods, and is also referred to as \textbf{score} in this study.
As the operational performance of the methods was also to be compared, the total time that each method took to run per instance was also measured.

Ideally, the heuristics could be compared by using the optimality gap, which is defined as the distance between the heuristic solution score and the optimal solution score.
However, as the randomly generated KEP instances of 300 nodes have shown to be intractable while using the PyCSP3 solver, i.e. impossible to solve optimally in a reasonable time, the optimality gap could not be measured.

For the trainable heuristics, the total training time was also measured and compared.
To monitor and guide the training process, the evolution of the loss function value over the training time was also collected.
Additionally, the mean score of the current model in the validation dataset was measured at each validation phase.

For a fair comparison between methods, all the measurements were made in the same test dataset, which is described in section \ref{dataset}, with the exception of the exact solution method, as explained in Section \ref{experiments} below.

\subsection{Solver Execution Time Analysis}
\label{solver_time_experiment}



To measure how much time it takes to solve a KEP instance in relation to the input size, the following experiment was designed:
first, we randomly generate 100 instances of each graph size (i.e. number of nodes), starting from 5 nodes and up to 300 nodes;
then, each one is solved optimally using the integer programming method described at section \ref{integer_programming}.
The elapsed time of each solver execution is collected for later analysis.

\subsection{Training of the ML Models}
\label{training_experiment}

The training of the ML models was separated in epochs.
In each epoch, we iterate through the 10 thousand instances of the train dataset, predicting, calculating the loss, and updating the GNN weights according to it.
At every 500 instances, a validation phase is run,
where the model is evaluated on the validation dataset and a checkpoint is saved.
The chosen batch size was 1, which means that the predictions were done on one instance at a time, because it empirically seemed to be the best for the learning process.
There were two machine learning models that were trained in this study:
the unconstrained unsupervised GNN and the two stage method, which are described at Section \ref{learnable_heuristics}.
For each training process, it was measured how the loss value evolved over time in the training and in the validation datasets. 

\subsection{Evaluation of KEP Solving Methods}
\label{methods_evaluation_experiment}

In order to do a fair comparison between the different heuristic methods, all of them were evaluated by predicting the solutions of all 10 thousand instances of the test dataset.
We did not set the cycles and paths size limit (parameter $k$ in constraint represented by Eq. \ref{eq:opt-constraint-cycle-size}), i.e. the cycles and paths could have any length, as long as they respect the KEP constraints.
We intend to assess the performance of these methods with limited cycles and paths length soon.
For each prediction, it was measured the solution score, the number of edges in the solution, and the relation between the solution score (which is the sum of the edge weights of the solution edges) and the total sum of edge weights of the graph.
In addition, the validity of the predicted solution was evaluated;
if invalid, we measure the number of invalid edges in the solution, i.e. the number of edges that disrespect the restrictions.
Furthermore, we measured the time it took for each method to solve each instance using a CPU.
Then, it was measured, for each model in relation to the whole test dataset, the mean, standard deviation, and distribution for the scores and the prediction elapsed times.

\section{Results}

\subsection{Solver Time Measurements}
\label{solver_time_results}

\begin{figure}
    \centerline{\includegraphics[width=95mm]{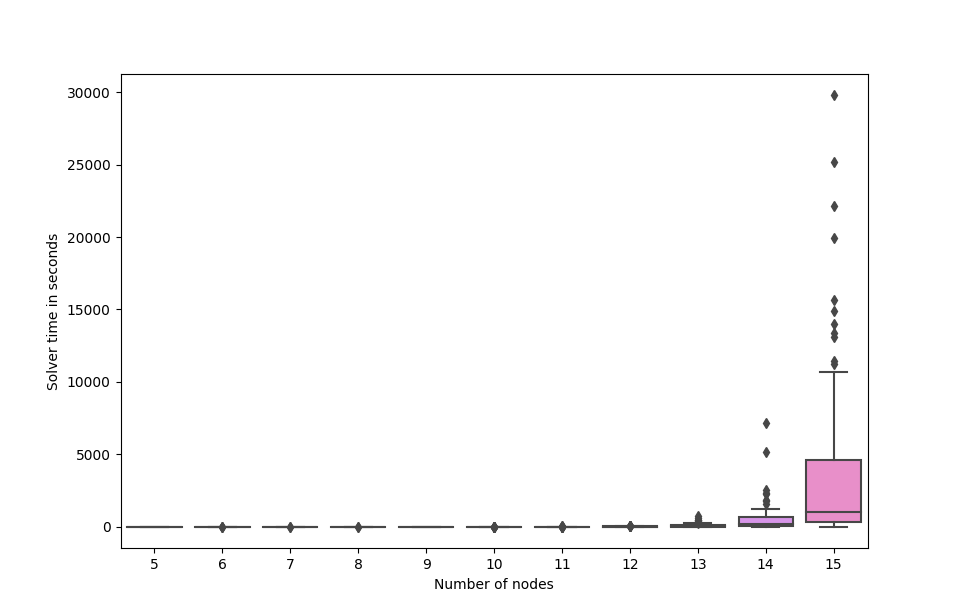}}
    \vspace{5mm}
     \caption{Boxplot of the time it takes to run the solver on KEP instances of sizes 5 to 15 (i.e. number of nodes).}
    \label{boxplot_solver_time}
\end{figure}


Figure \ref{boxplot_solver_time} shows a box plot of the time that the solver took to optimally solve KEP instances in relation to the instance size.
For that, graphs of sizes 5 to 15 (i.e. number of nodes) were used;
initially, graphs with up to 300 nodes were going to be included in the analysis, but as solving graphs with 16 nodes or more would take several days to compute, they were excluded.
To solve a hundred instances with 15 nodes, for instance, it took 101.2 hours in total.

As we can see, the experiment results show a pattern of exponential growth of computational time in relation to the input size. 
The mean time it took when the graph node number was beneath 10 was always below 1.5 seconds. 
For instances with 15 nodes, the mean time measured was 3569.33 seconds, i.e. roughly one hour.

\subsection{Training of the ML Models}
\label{training_results}

\begin{figure}
    \centerline{\includegraphics[width=85mm]{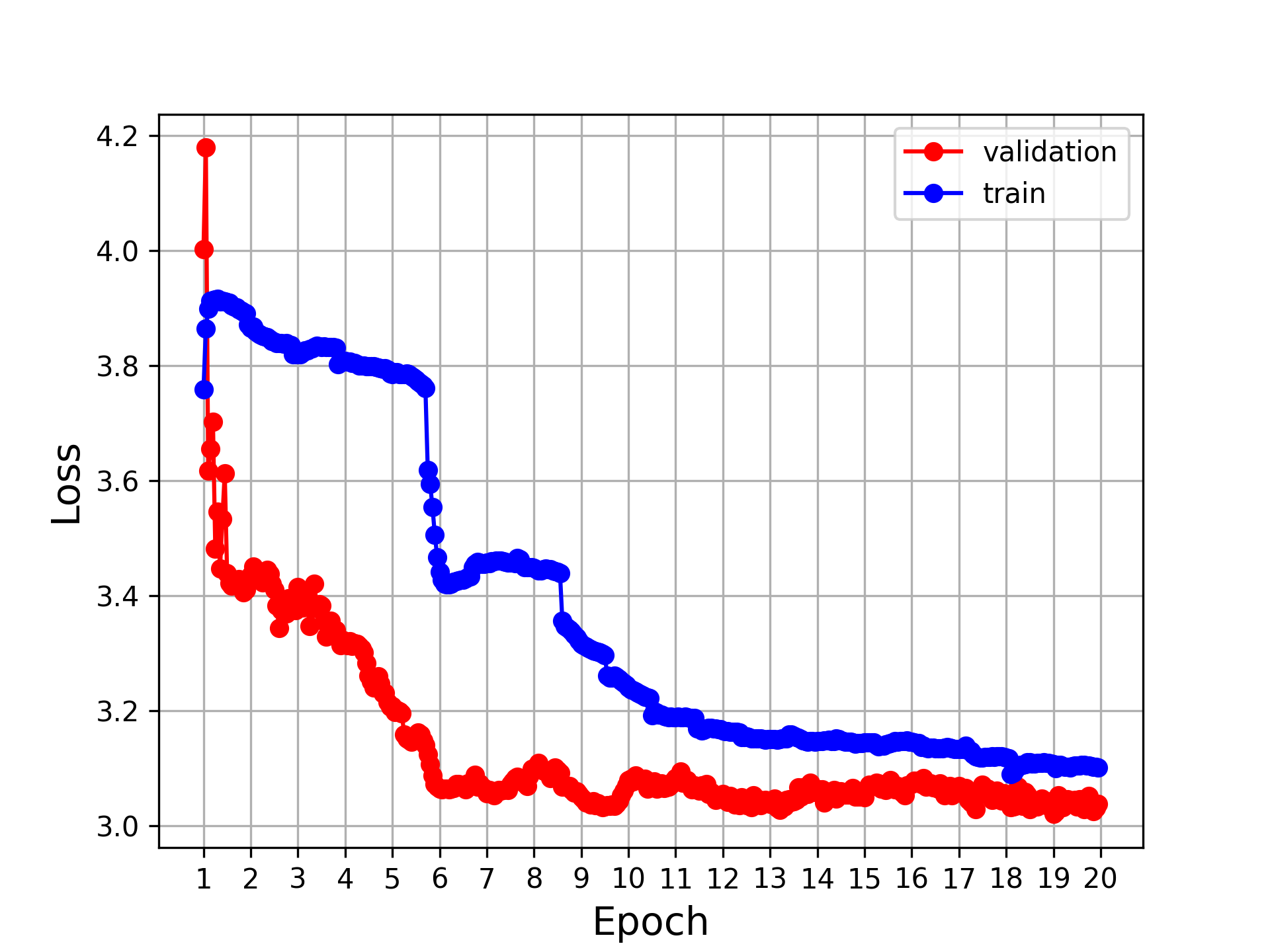}}
    \vspace{5mm}
     \caption{Evolution of the training and validation loss for \textit{GNN+GreedyPaths} method.}
    \label{GNN_paths_mean_loss_plot}
\end{figure}

\begin{figure}
    \centerline{\includegraphics[width=85mm]{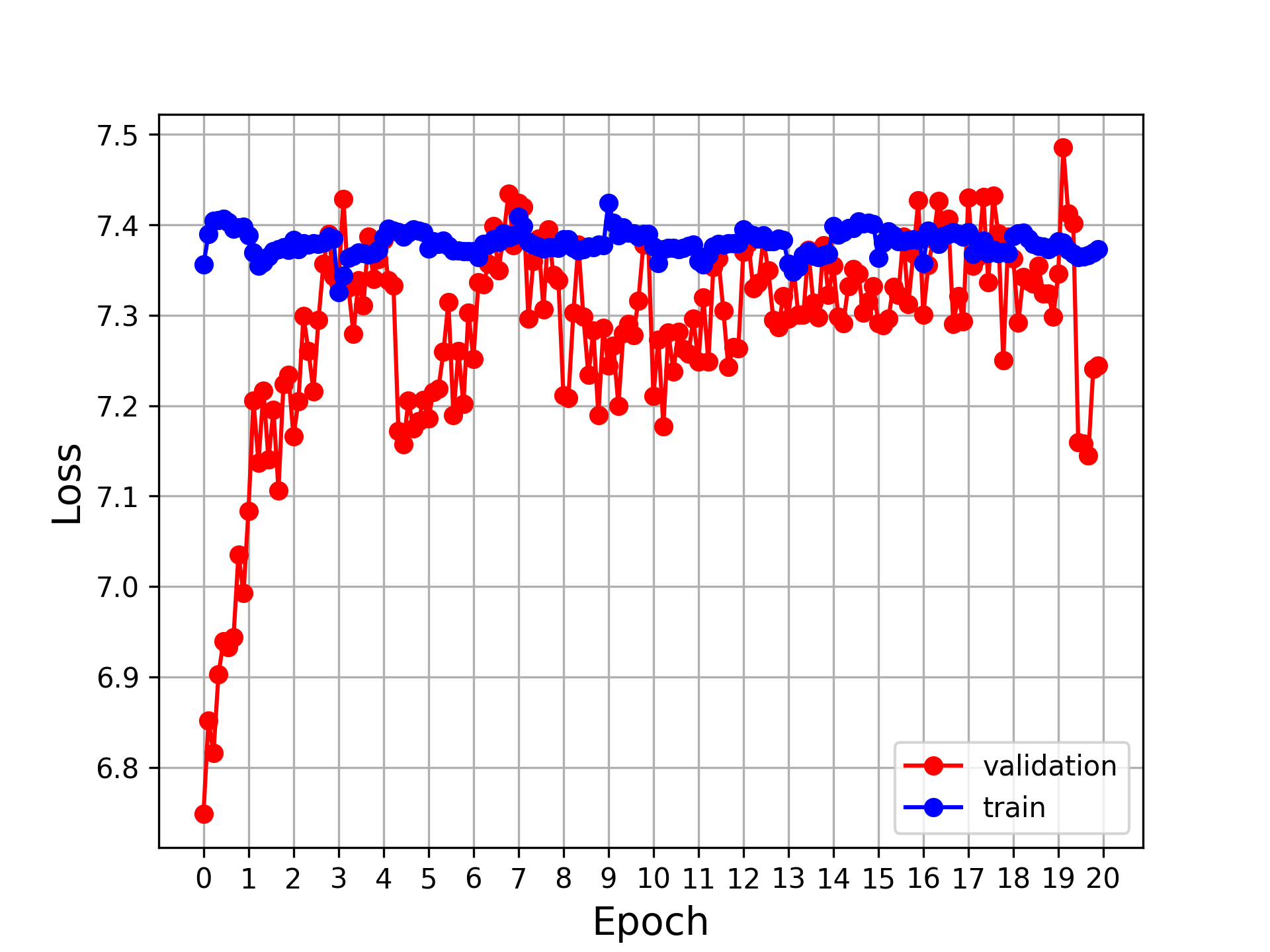}}
    \vspace{5mm}
     \caption{Evolution of the training and validation loss for \textit{GNN+GreedyCycles} method.}
    \label{GNN_cycles_mean_loss_plot}
\end{figure}

\begin{figure}
    \centerline{\includegraphics[width=85mm]{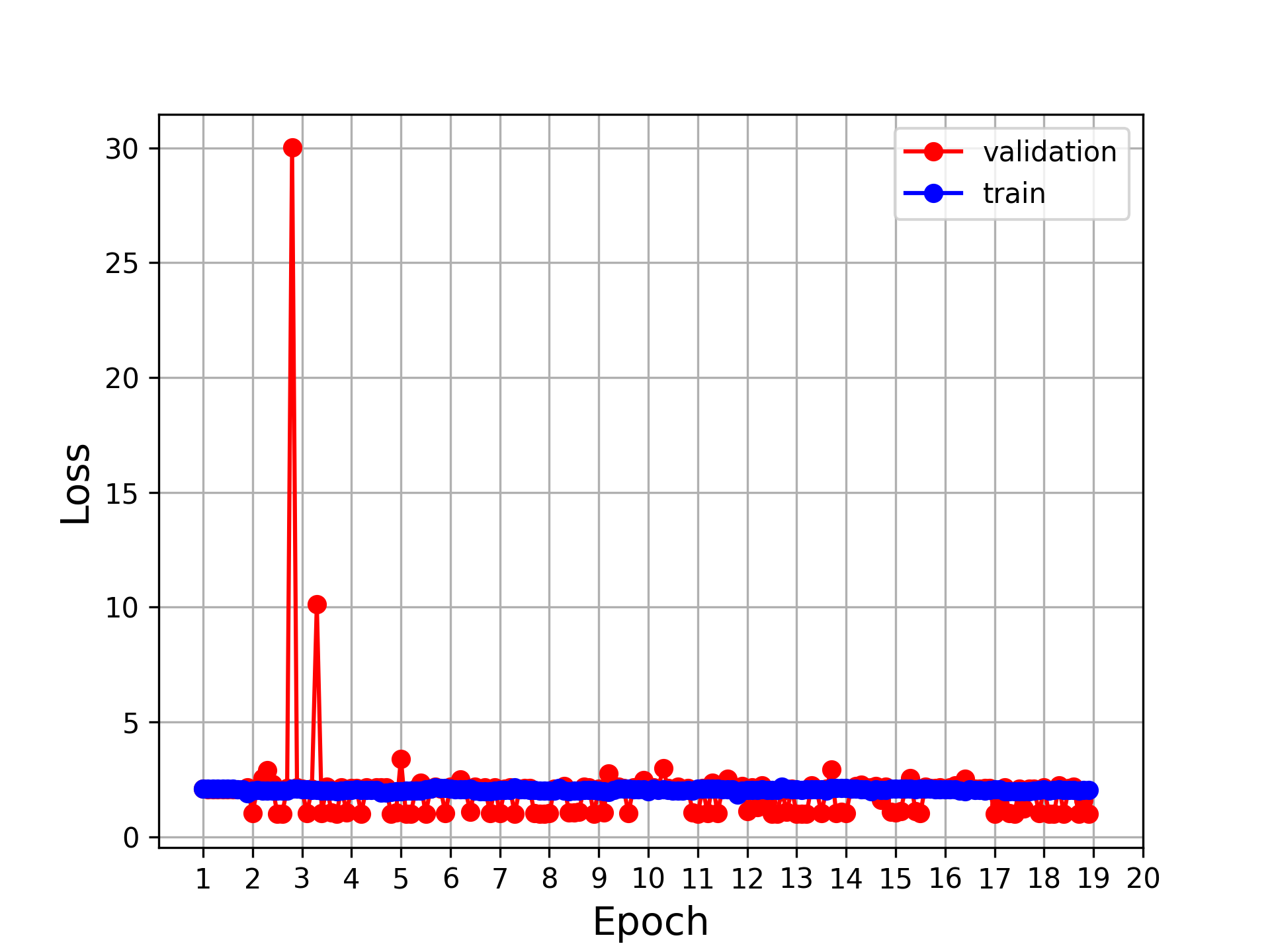}}
    \vspace{5mm}
     \caption{Evolution of the training and validation loss for the \textit{Unsupervised GNN} method.}
    \label{unsupervised_GNN_mean_loss_plot}
\end{figure}


\begin{figure}
    \centerline{\includegraphics[width=85mm]{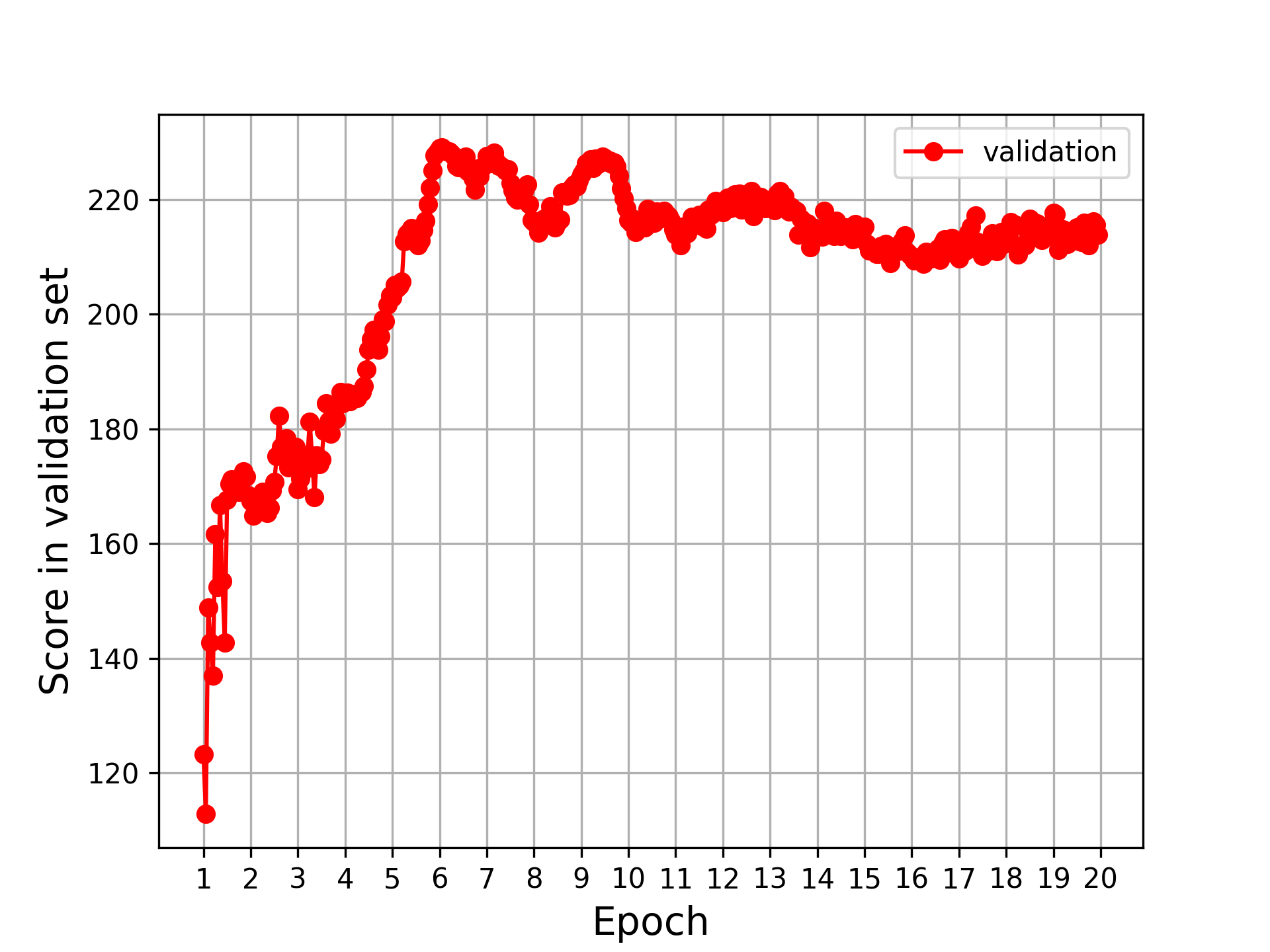}}
    \vspace{5mm}
     \caption{Evolution of the score measured in the validation dataset for the \textit{GNN+GreedyPaths} method.}
    \label{GNN_paths_mean_score_plot}
\end{figure}

\begin{figure}
    \centerline{\includegraphics[width=85mm]{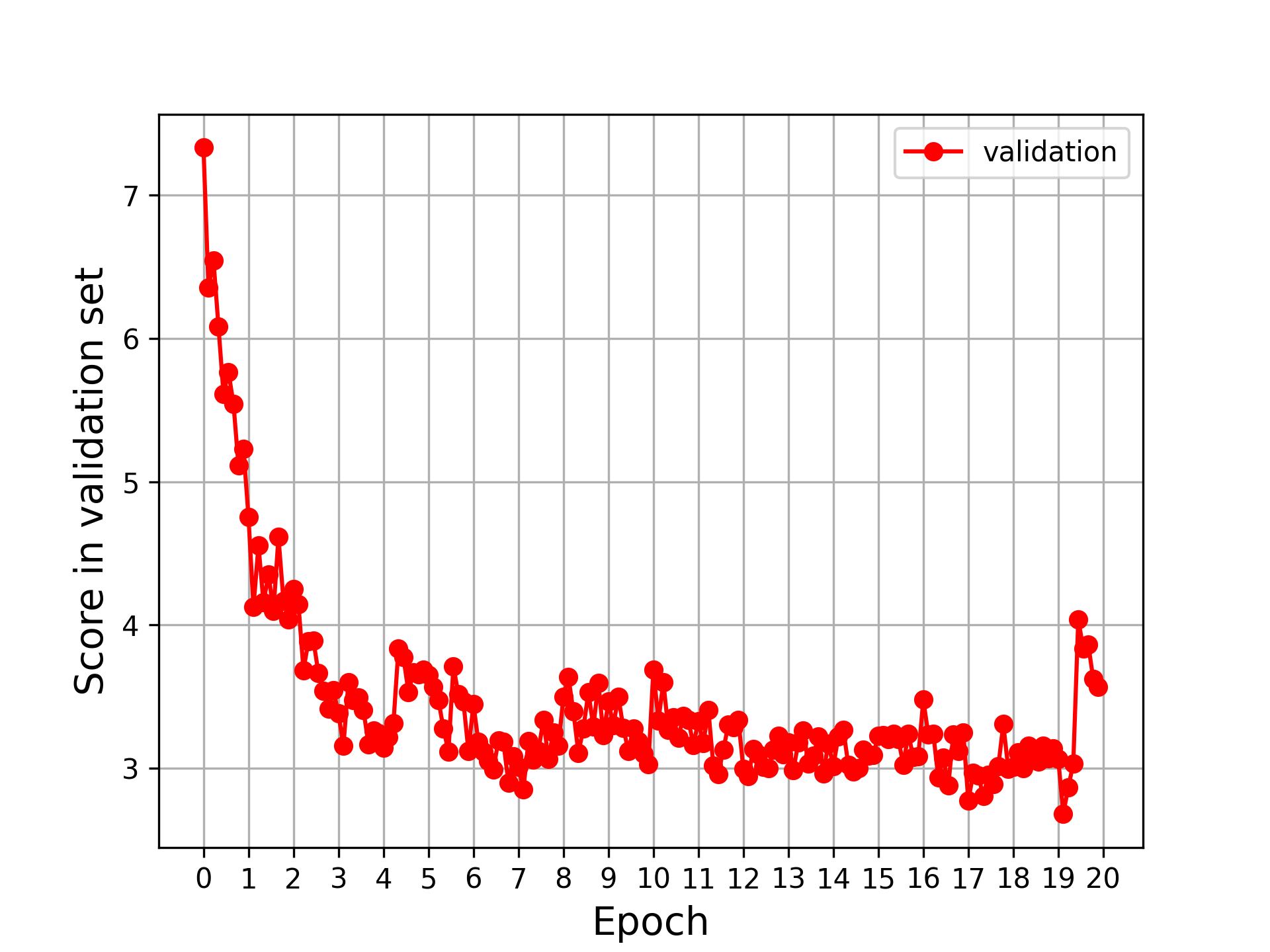}}
    \vspace{5mm}
     \caption{Evolution of the score measured in the validation dataset for the \textit{GNN+GreedyCycles} method.}
    \label{GNN_cycles_mean_score_plot}
\end{figure}

\begin{figure}
    \centerline{\includegraphics[width=85mm]{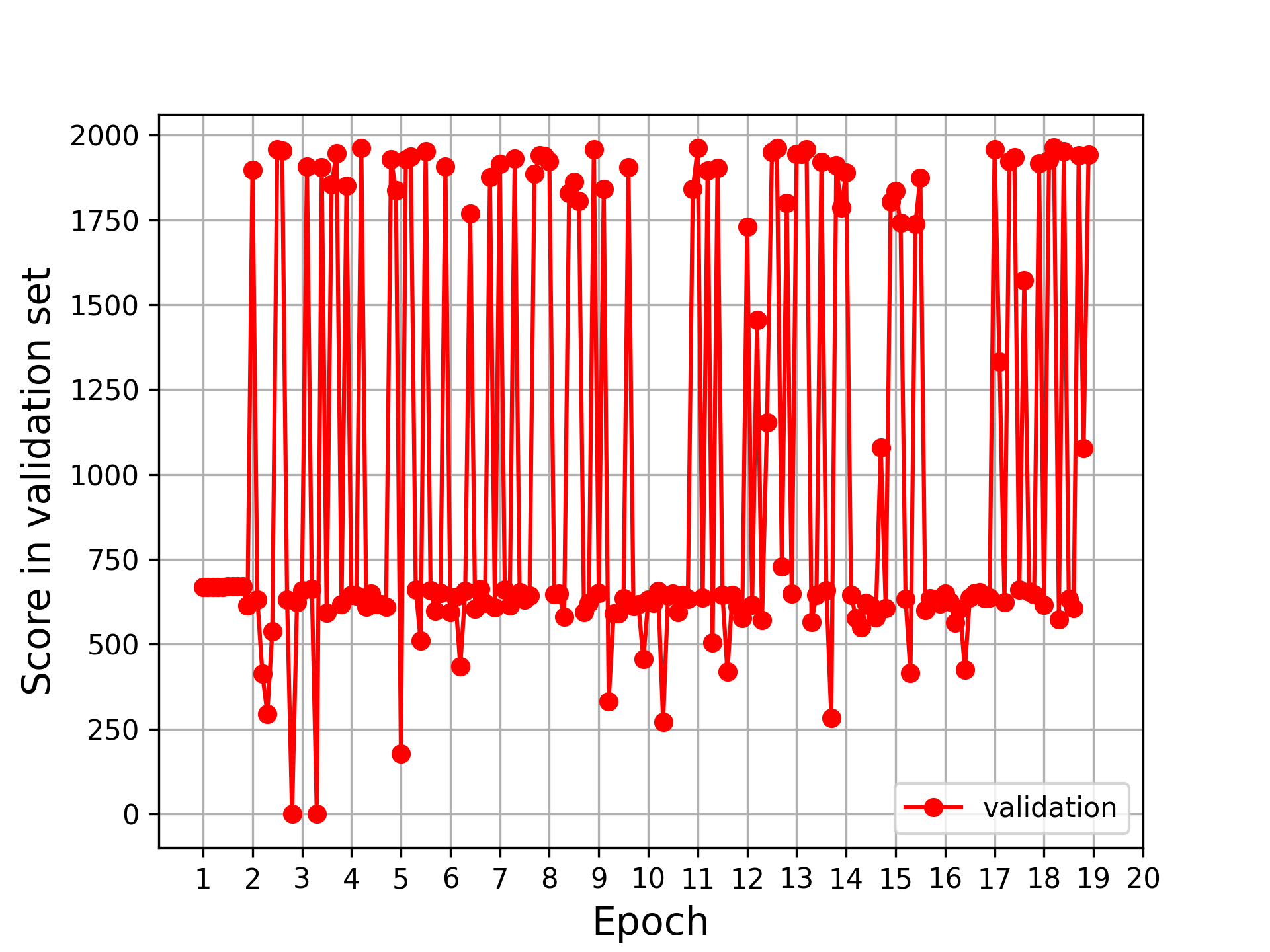}}
    \vspace{5mm}
     \caption{Evolution of the score measured in the validation dataset for the \textit{Unsupervised GNN} method.}
    \label{unsupervised_GNN_mean_score_plot}
\end{figure}

\begin{figure}
    \centerline{\includegraphics[width=85mm]{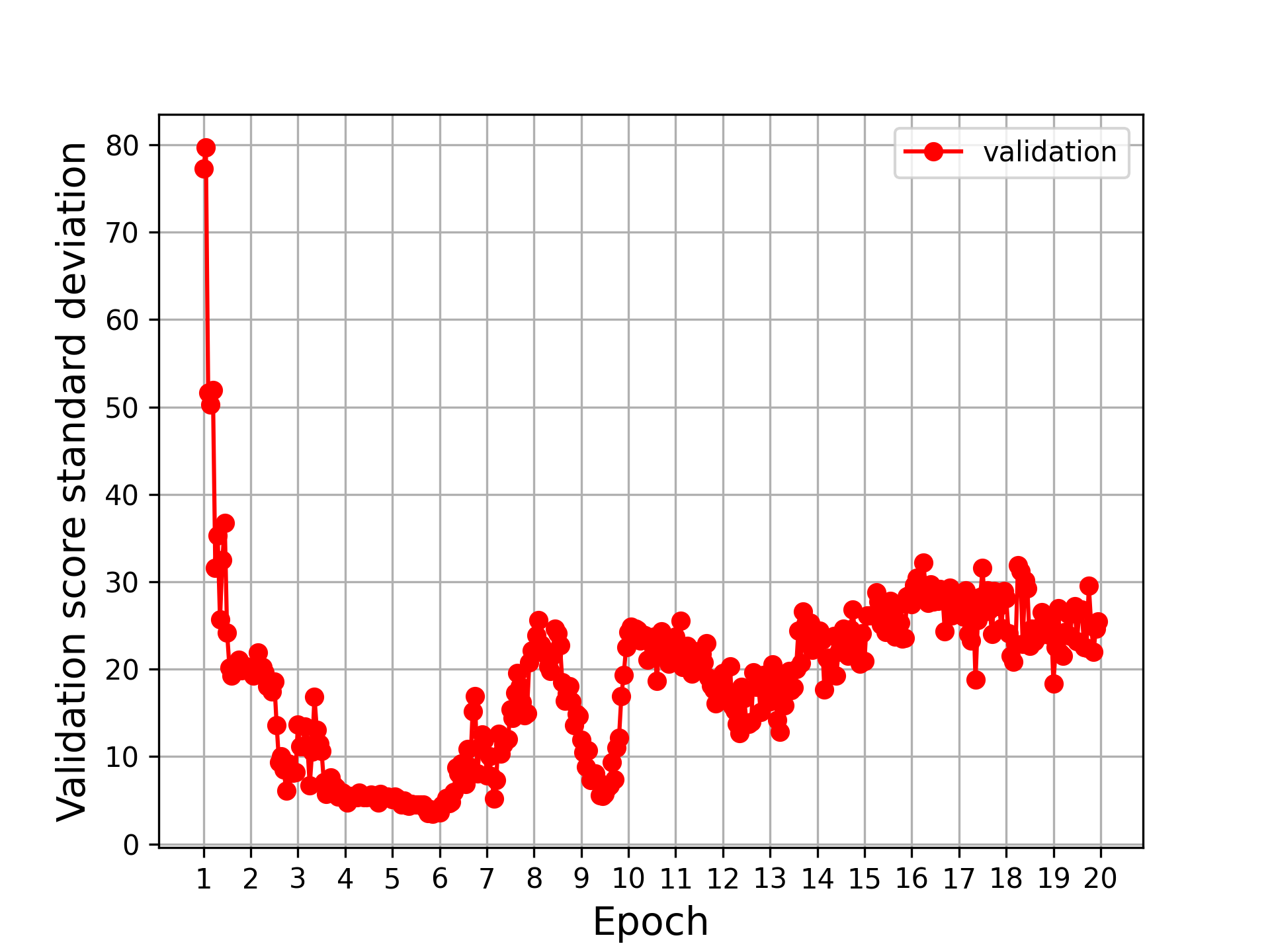}}
    \vspace{5mm}
     \caption{Evolution of the standard deviation of score measured in the validation dataset for the \textit{GNN+GreedyPaths} method.}
    \label{GNN_paths_std_score_plot}
\end{figure}

Figures \ref{GNN_paths_mean_loss_plot}, \ref{GNN_cycles_mean_loss_plot}, and \ref{unsupervised_GNN_mean_loss_plot} show the evolution of the loss value on the training and validation datasets for the two stage methods described at Subsection \ref{two_stage_method}, \textit{GNN+GreedyPaths} and \textit{GNN+GreedyCycles}, and for the \textit{Unsupervised GNN} method described at Subsection \ref{unconstrained_gnn_unsupervised}, respectively.

Figures \ref{GNN_paths_mean_score_plot}, \ref{GNN_cycles_mean_score_plot}, and \ref{unsupervised_GNN_mean_score_plot} show the evolution of the the mean score value for \textit{GNN+GreedyPaths}, \textit{GNN+GreedyCycles}, and \textit{Unsupervised GNN}, respectively.
Figure \ref{GNN_paths_std_score_plot} shows the evolution of the standard deviation of the scores predicted on the validation dataset.

\subsection{Methods' Performances}
\label{methods_performances_result}

\begin{figure}
    \centerline{\includegraphics[width=85mm]{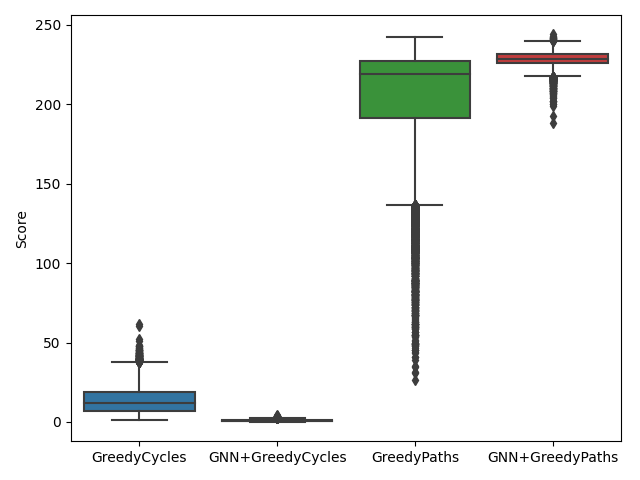}}
    \vspace{5mm}
     \caption{Box plot comparing the approximate solution scores obtained when each of the evaluated methods was used in the test dataset.
     The evaluated methods were two non-learnt heuristics, \textit{GreedyCycles} and \textit{GreedyPaths}, and their 2 stage method versions, \textit{GNN+GreedyCycles} and \textit{GNN+GreedyPaths}.}
    \label{boxplot_method_score}
\end{figure}

Figure \ref{boxplot_method_score} shows a box plot of the approximate solution scores (i.e. the sum of the weights of the edges contained in the approximate solution) achieved by each method in the test dataset, with the exception of \textit{Unsupervised GNN} and \textit{integer programming}.
As all the solutions found by the \textit{Unsupervised GNN} method were invalid, there was no reason to evaluate their quality.
The integer programming method is also absent from the plot, as it was not possible to run it in instances with 300 nodes.

As we can see, the GreedyPaths heuristic method found much better solutions than GreedyCycles.
The 2 stage method variations, \textit{GNN+GreedyCycles} and \textit{GNN+ GreedyPaths}, obtained very different performances.
Unfortunately, the GNN module in the \textit{GNN+GreedyCycles} method did not manage to learn to output edge scores that help the heuristic;
on the contrary, the quality of its approximate solutions are considerably worse than those from \textit{GreedyCycles}.
The \textit{GNN+GreedyPaths} method, however, effectively learned to output better solutions than the non-learnt heuristic achieving a mean score of 228.40 on the test dataset, while \textit{GreedyPaths} obtained 203.79; this shows an improvement of 12\% of the mean solution score with the use of the GNN.
We can also see that, although the best scores achieved by each of them are very similar, the score distribution is very different:
while \textit{GreedyPaths} outputs approximate solutions with very large range of scores, \textit{GNN+GreedyPaths}'s approximate solutions have much more consistent scores, obtaining a decent performance throughout all instances of the dataset.

\subsection{Methods' Computational Time}
\label{prediction_time_results}

\begin{figure}
    \centerline{\includegraphics[width=92mm]{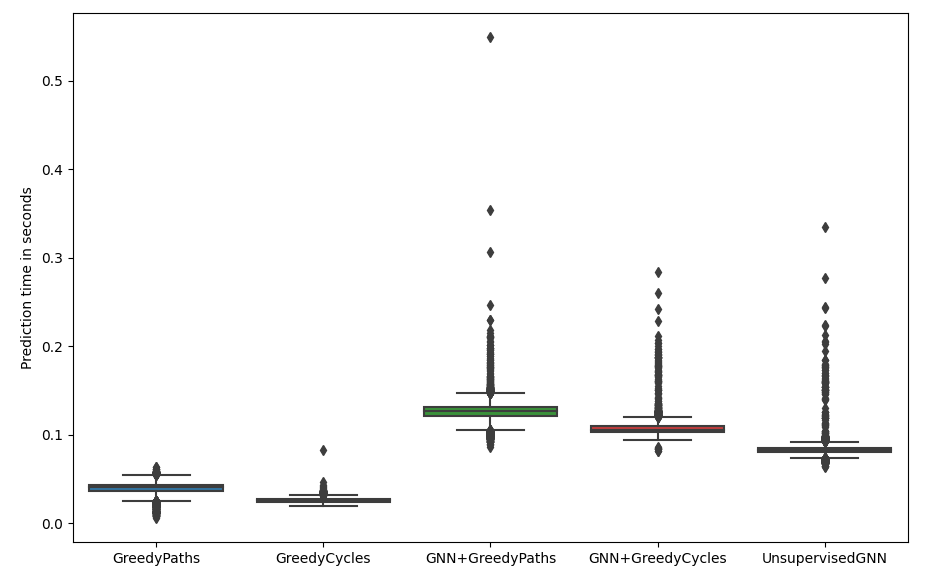}}
    \vspace{5mm}
     \caption{Box plot comparing the time to compute a solution on each of the 10 thousand KEP instances of the test dataset, each one with 300 nodes.
     The evaluated methods were two non-learnt heuristics, \textit{GreedyCycles} and \textit{GreedyPaths}, their 2 stage method versions, \textit{GNN+GreedyCycles} and \textit{GNN+GreedyPaths}, and \textit{UnsupervisedGNN}, which is a GNN trained and used without an heuristic.}
    \label{boxplot_prediction_time}
\end{figure}

The box plot in Figure \ref{boxplot_prediction_time} shows the comparison between the time it took for each method to solve the KEP instances of the test dataset.
As we can see, all of them took less then a second.
Although the difference is not large, the two basic search heuristics took less time than the GNN based methods, and \textit{GNN+GreedyPaths} took, on average, the most time.

\section{Analysis of Experimental Results}
\label{discussion}

\subsection{Solver Time Analysis}

Because the Kidney Exchange problem is NP Hard, the time it takes to optimally solve each instance is expected to grow exponentially as the instance size grows.
Regardless, considering that in \cite{KEP_2015} real life KEP instances could be optimally solved in a reasonable time, it was expected that we would also be able to optimally solve the ones used in this study, since they have been constructed to have similar sizes to the ones used in the article.
However, as described in Subsection \ref{solver_time_results}, instances of size as small as 15 already took in average 1 hour to solve.
Considering that we would want to optimally solve all 10 thousand instances of the test set in order to fairly compare to the other methods and to measure their optimality gap, this process would take an unreasonable time, estimated to be around 10 thousand hours, i.e. roughly 1.14 years.
This estimate is only if the KEP instances on the test dataset had 15 nodes;
for instances with 300 nodes, it would surely take an unreasonably enormous amount of time.

The number of nodes of the input instance is not, however, the sole factor that determines the time it takes to optimally solve it;
in a set of instances with the same number of nodes, some are "harder" than others, i.e. take more time to solve.
As the instance size increases, so does the variability of the time to solve it:
the minimum and maximum times measured for instances with 15 nodes were 5.17 seconds and 29779.01 seconds (i.e. roughly 8 hours);
for comparison, the minimum and maximum times for instances with 5 nodes were 0.8 and 1.3 seconds.
Hence, it is possible that, while some real life instances are solvable in a reasonable time, a percentage of them would take too long to solve, thus becoming intractable.

There can be several reasons why the authors of \cite{KEP_2015} could compute the optimal solution of their KEP instances in much less time.
First of all, they probably used a solver tool that is much more efficient than PyCSP3.
In addition, the computer used may be much more powerful than the one used in this study.
Also, it is important to remember that the NP-Hardness of KEP guarantees that the computational time it takes to solve the worst case scenario grows exponentially, but in practice real life instances may often have specific properties which may cause them to be either harder or easier to solve.
Hence, another plausible reason is that their instances may be much easier to solve.
Furthermore, the authors used a constrain relaxation technique that speeds up significantly the solving process.
These set of reasons alone may not explain totally why they were able to optimally solve their KEP instances much faster than we did on our data;
this may be further investigated in future work.

\subsection{Training of the GNN model}

As we can see in Figure \ref{GNN_paths_mean_loss_plot}, the training of the \textit{GNN+GreedyPaths} method was successful, seen as it managed to optimize the GNN by minimizing the loss function.
The training of the \textit{GNN+GreedyCycles} and \textit{Unsupervised GNN} methods, however, were unsuccessful, as shown by the loss curves of Figures \ref{GNN_cycles_mean_loss_plot} and \ref{unsupervised_GNN_mean_loss_plot}, which do not decrease over time.

Although at first glance at Figure \ref{unsupervised_GNN_mean_score_plot} the \textit{Unsupervised GNN} model seems to achieve great scores, unfortunately all its output solutions were invalid, i.e. they did not comply to the KEP constraints.
It is clear that the loss constraint regularization (described at \ref{kep_loss_constraint_regularization}) added to the loss function was unsuccessful in helping the model learn to comply to the KEP constraints.
This highlights the necessity of having a methods that guarantee with total certainty that all its output solutions are valid.
However, even though there were many manual trials with different hyperparameter combinations, it is still possible that a variation of this technique could work with a different setting, i.e. another GNN architecture, other hyperparameters, and so on.

Because of the skip connection that sums the original edge weights to the predicted edge scores at the end of the GNN, the predicted solutions start off very similar to the ones made by \textit{GreedyPaths}.
Then, the changing of the GNN weights disrupts these scores, which increases the loss, but goes on to improve them, eventually arriving at a performance that is better than \textit{GreedyPaths}.
After some point (around epoch 6 in Figure \ref{GNN_paths_mean_score_plot}), the learning converges to a solution, and after a while the performance starts to slowly worsen.
The final model was chosen from the checkpoint with the highest score measured on the validation dataset, which was in epoch 6, step 3500.
As we can see on Figure \ref{GNN_paths_std_score_plot}, at this point the model's scores on the validation also presented the lowest standard deviation, which indicates that the model's predictions were more consistent, maintaining a decent performance throughout all instances.

\subsection{Methods' Performances}
\label{methods_performance_discussion}

Ideally, we would want to evaluate and compare each method by measuring their optimality gap for each instance, i.e. how far the approximate solution is from the optimal one.
However, as we do not have access to the optimal solution, this was not possible.
We can nevertheless estimate it roughly by examining an upper bound:
each instance has 300 nodes, and each node may donate and receive at most one kidney;
thus, the solution with the most edges would be contain cycles that together comprehend all nodes.
As each edge weight is a value between 0 and 1, the maximum score possible is equal to 300, when all edges in the cycle have a weight of 1.
Hence, an upper bound for the score is the number of nodes, which in this case is 300.
This is obviously extremely unrealistic, as it assumes that all nodes are PDPs, that there are a set of cycles that links all of them, and that the solution edge weights are equal to 1
(as the edge weights are values sampled from a uniform distribution between 0 and 1, their average value is 0.5).
Considering the score upper bound of 300 as a very conservative estimate for the mean optimal solution value, we can estimate that the absolute and relative optimality gap for the \textit{GreedyPaths} method would be 96.28 and 32\%;
for \textit{GreedyCycles}, these values would be 286.29 and 95.4\%;
for \textit{GNN+GreedyPaths}, 71.59 and 23.8\%;
for \textit{GNN+GreedyCycles}, 299.15 and 99.7\%.
Hence, the improvement on the optimality gap of the \textit{GNN+GreedyPaths} method in relation to \textit{GreedyPaths} would be at least 34.4\% (96.28 to 71.59), which is already a very substantial improvement.


It is clear that the two methods that searched for paths performed much better than the ones that searched for cycles.
There are many possible explanations for this observed behaviour.
Maybe the best cycles-only solution in a KEP instance is usually much worse than the best paths-only solution.
This, however, can only be verified by making comparisons to the cycles-only and paths-only exact solutions, which are unavailable.
Also, the \textit{GreedyCycles} method is probably less efficient because it discards the path it is constructing if it does not end up closing a cycle.
It also shortens the constructed cycle if it closes before the node it had begun on, which may also lead to worse performance overall.
The \textit{GreedyPaths} method does not have these issues, as it always keeps the edges it adds to each path it constructs.

We can also observe that the while the GNN module in the \textit{GNN+GreedyPaths} improved the performance in relation to the basic non-learnable heuristic, in \textit{GNN+GreedyCycles} it only worsened it.
It is possible that its GNN module in \textit{GNN+GreedyCycles} could not learn the needed context to know if a given edge would lead to a longer and higher-valued cycle because it is too complex, and does not depend that much on the 3-neighborhood context, which is the limit of information gather in each node with the GNN architecture used, as it only has 3 message passing GNN layers.
Another possible explanation is that it is way harder for a model to learn to compute edge scores that help the choices of \textit{GreedyCycles} because it is inherently more complex than \textit{GreedyPaths}, i.e. it is not just a sequence of simple decisions, as it also has to keep track of the rest of the nodes of the cycle being constructed, check if it closed a cycle, and remove from the solution in construction the edges added before the node where the cycle was closed.
Put simply, the more complex the second step heuristic is, the harder it is for a machine learning model to learn to help it.

As explained at Section \ref{methods_performances_result}, \textit{GNN+GreedyPaths} approximate solutions have much more consistent scores, which suggests that it probably handle much better "hard" instances.
A plausible interpretation is that the GNN module helps the subsequent greedy heuristic to avoid choosing edges that are only locally good, but lead to worse paths overall.
It is able to do this because it considers information of the neighborhood context.

\subsection{Methods' Computational Time}

The GNN computational complexity is linear, as it performs a fixed amount of computations per graph node plus another fixed amount of computations per edge.
As for the heuristic methods \textit{GreedyPaths} and \textit{GreedyCycles}, their computational complexity is also linear, as the worst case scenario one edge for each node will be added, one by one, into the solution;
hence, it always performs a quantity of computational operations linearly proportional to the number of nodes of the input instance, at worst.

The results from Subsection \ref{prediction_time_results} show that every method tested in this work took very little time to execute,
with the exception of the \textit{integer programming} method, which took so much time that applying it to 300 nodes instances became intractable.
The \textit{GreedyPaths} method took a bit more time than \textit{GreedyCycles} probably because it found better solutions overall, and consequently took more computing steps to construct each solution.
The same effect may also explain the prediction time difference between \textit{GNN+GreedyPaths} and \textit{GNN+GreedyCycles}.
The \textit{UnsupervisedGNN} method took a bit less time to execute than the two step methods, which was expected because it runs the same computations, but without the second step, which is the basic search heuristic.

\section{Conclusion and Future Work}
\label{conclusion}

In this work, several heuristic methods with and without machine learning for approximately solving the Kidney Exchange Problem were proposed and investigated.
They were tested on an artificial dataset and compared between each other and with an implementation of an exact solution method.
Additionally, it was made an experiment for measuring the time it took for the exact solution method to solve an instance in relation to the instance size.
The results of the evaluations and experiments were then analysed and discussed.

\subsection{Answers to the Research Questions}

As seen from the results in subsection \ref{methods_performance_discussion}, the main question presented at Section \ref{introduction} was answered: \textbf{yes, the Kidney-Exchange problem can be better approximately solved with the help of machine learning}.

As for the feasibility of the ML methods, the \textit{GNN+GreedyPaths} method surpassed all other evaluated heuristics in terms of the quality of the solutions it provides;
among all evaluated methods, it remains the one that best approximately solves the dataset instances in a reasonable time.
The other ML methods evaluated in the work (\textit{UnsupervisedGNN} and \textit{GNN+GreedyCycles}), however, did not achieve good results.

Regarding the viability of such methods in terms of computational time, the GNN module adds an almost insignificant overhead when compared to the non-learnable heuristics.
When compared to the solver running the exact solution, it is several orders of magnitude faster for instances with at least 15 nodes.
The complexity of the two stage methods is linear, turning instances that were previously intractable due to their size into easily approximately solvable in a reasonable time.

As for the limitations of the employed machine learning methods for this problem, it is clear that they are not simple to use, as they need to be properly trained, which is not easy to do.
Although using the two stage method potentially improves considerably the performance in relation to the basic heuristics, as happened with the \textit{GNN+GreedyPaths} method, it also introduces several new hyperparameters which need to be adequately set in order for the method to work.
Applying supervised learning turned out to be unfeasible because of the need for the exact solution to be used as edge labels.
Regarding the \textit{UnsupervisedGNN} method, our results suggest that imposing the constraints through adding terms in a loss function is actually really hard;
hence, the method never learns to output valid solutions, rendering it useless.

\subsection{Main Contributions}

In the following list, a summary of each of the main contributions that this work provided is presented.

    
    \textbf{Learnable Heuristics for KEP:}
    Although the two stage approach was already used by past work \cite{gnn-beam_search_2019, graph_learning_comb_opt_survey_2020, learning_tsp_2022}, this was the first work to adapt it and apply it to KEP.
    Two variations of the approach were implemented: \textit{GNN+GreedyPaths} and \textit{GNN+GreedyCycles}, the first one having achieved satisfactory performance.
    
    \textbf{Non-Learnable Heuristics for KEP:}
    Two new deterministic heuristic methods for approximately solving the Kidney Exchange Problem were introduced: \textit{GreedyPaths} and \textit{GreedyCycles}.
    They were also evaluated in the test dataset, giving insight on their effectiveness.
    
    \textbf{Node-wise Softmax:}
    A variation of the softmax activation function designed to be applied to edge scores in graph problems was created and implemented.
    It showed to be useful for the GNN, empirically improving the its performance.
    The author plans to contribute to the PyTorch library with the implementation of this technique, thus making it available and easily usable by its future users.
    
    \textbf{KEP Unsupervised Loss:}
    A novel loss function (described at \ref{kep_unsupervised_loss}) designed for KEP was introduced.
    It optimizes the weighted sum of the edges in the predicted solution without the need for supervision.
    It was validated in the training of the \textit{GNN+GreedyPaths} method, as seen in Figure \ref{GNN_paths_mean_loss_plot} in Section \ref{training_results}, where it lead the GNN to learn to effectively help the heuristic method construct better approximate solutions.
    
    

\subsection{Future Directions}

    \textbf{Using real data:}
    Use data collected by countries' or hospital's healthcare system to evaluate how the presented methods would perform in real life situations.
    This would also allow us to compare our proposed methods with others that were already evaluated in the same data.
    
    \textbf{Using better artificial data:}
    There more sophisticated methods for generating artificial KEP instances, such as the ones presented at \cite{saidman_2006} and \cite{improved_kep_instance_generation_2022}.
    They are still far from sufficiently similar to real data so as to substitute evaluating it. However, it would still probably give an evaluation of KEP solving methods that is closer to that of real life situations.
    Furthermore, training the model with these instances could also potentially lead to better results.
    Another possibility of generating better artificial data would be to use a graph generator model that learns to create instances similar to the real data; the Graph Variational Auto-Encoder presented at \cite{graph_vae} and the MolGAN presented at \cite{molgan} are good examples of candidate methods to be adapted for that goal.
    
    
    
    \textbf{Training the models with supervised learning:}
    Another direction is to develop a method that learns with supervision, evaluate it, and compare it to the other methods.
    For that, we would first have to obtain the optimal solutions, which would then be used as labels for the supervised training.
    This could be done either by improving a lot the integer programming method's speed and/or by training on much smaller instances (i.e. instances with less than 15 nodes).
    Another promissing variation of this idea is to use the N best solutions, and create soft labels where each edge would have a value between 0 and 1 that would indicate how often it appears in the best solutions, weighted by the quality of these solutions.
    \textbf{Running a GNN before every step of the greedy heuristic:}
    Instead of running the GNN once and then passing the edge scores to the greedy heuristic method, new node embeddings could be generated before each step of the heuristic.
    Although significantly costlier in terms of inference, as the GNN is executed many times per instance, this has shown good results for other graph route optimization problems \cite{attention_learn_tsp_2018}.
    This approach is called \textit{autoregressive decoding} and explored for solving the TSP by \cite{learning_tsp_2022}.
    



    

\section*{Acknowledgments}
This work is supported in part by the Brazilian Research Council CNPq and the CAPES Foundation - Finance Code 001.

\bibliographystyle{ieeetr}
\bibliography{kep_gnn}

\begin{thebibliography}{10}

\bibitem{kidney_exchange_2004}
A.~Roth, T.~Sönmez, and U.~Unver, ``Kidney exchange,'' {\em The Quarterly
  Journal of Economics}, vol.~119, no.~2, pp.~457--488, 2004.

\bibitem{KEP_2015}
R.~Anderson, I.~Ashlagi, D.~Gamarnik, and A.~E. Roth, ``Finding long chains in
  kidney exchange using the traveling salesman problem,'' {\em Proceedings of
  the National Academy of Sciences}, vol.~112, no.~3, pp.~663--668, 2015.

\bibitem{improved_kep_instance_generation_2022}
M.~Delorme, S.~García, J.~Gondzio, J.~Kalcsics, D.~Manlove, W.~Pettersson, and
  J.~Trimble, ``Improved instance generation for kidney exchange programmes,''
  {\em Computers and Operations Research}, vol.~141, p.~105707, 2022.

\bibitem{economic_assessment_kidney_transplant_2018}
D.~A. Axelrod, M.~A. Schnitzler, H.~Xiao, W.~Irish, E.~Tuttle-Newhall, S.-H.
  Chang, B.~L. Kasiske, T.~Alhamad, and K.~L. Lentine, ``An economic assessment
  of contemporary kidney transplant practice,'' {\em American Journal of
  Transplantation}, vol.~18, no.~5, pp.~1168--1176, 2018.

\bibitem{longest_KEP_chain}
``Longest kidney transplant chain - guinness world record organization
  distinguishes the national kidney registry for world’s longest kidney
  transplant chain.''
  https://transplantsurgery.ucsf.edu/news--events/ucsf-news/88223/UCSF-Part-of-Longest-Kidney-Transplant-Chain---Guinness-World-Record-Organization-Distinguishes-the-National-Kidney-Registry-for-World\%E2\%80\%99s-Longest-Kidney-Transplant-Chain,
  dec 2020.
\newblock Accessed: 2023-03-01.

\bibitem{biro_2021}
P.~Biró, J.~{van de Klundert}, D.~Manlove, W.~Pettersson, T.~Andersson,
  L.~Burnapp, P.~Chromy, P.~Delgado, P.~Dworczak, B.~Haase, A.~Hemke,
  R.~Johnson, X.~Klimentova, D.~Kuypers, A.~{Nanni Costa}, B.~Smeulders,
  F.~Spieksma, M.~O. Valentín, and A.~Viana, ``Modelling and optimisation in
  european kidney exchange programmes,'' {\em European Journal of Operational
  Research}, vol.~291, no.~2, pp.~447--456, 2021.

\bibitem{abraham_2007}
D.~J. Abraham, A.~Blum, and T.~Sandholm, ``Clearing algorithms for barter
  exchange markets: Enabling nationwide kidney exchanges,'' in {\em Proceedings
  of the 8th ACM Conference on Electronic Commerce}, EC '07, (New York, NY,
  USA), p.~295–304, Association for Computing Machinery, 2007.

\bibitem{csc_tsp_ml_2020}
Y.~Yang and J.~Rajgopal, ``Learning combined set covering and traveling
  salesman problem,'' {\em arXiv preprint arXiv:2007.03203}, 2020.

\bibitem{graph_coloring_gnn_2019}
H.~Lemos, M.~O.~R. Prates, P.~H.~C. Avelar, and L.~C. Lamb, ``Graph colouring
  meets deep learning: Effective graph neural network models for combinatorial
  problems,'' in {\em 31st {IEEE} International Conference on Tools with
  Artificial Intelligence, {ICTAI} 2019, Portland, OR, USA, November 4-6,
  2019}, pp.~879--885, {IEEE}, 2019.

\bibitem{graph_coloring_gnn_2020}
H.~L.~d. Santos, ``Solving the decision version of the graph coloring problem:
  a neural-symbolic approach using graph neural networks,'' Master's thesis,
  UFRGS {F}ederal {U}niversity, {P}orto {A}legre, {B}razil, 2020.

\bibitem{gnn_minimum_vertex_cover_2019}
R.~Sato, M.~Yamada, and H.~Kashima, ``Approximation ratios of graph neural
  networks for combinatorial problems,'' in {\em Advances in Neural Information
  Processing Systems 32} (H.~M. Wallach, H.~Larochelle, A.~Beygelzimer,
  F.~d'Alch{\'{e}}{-}Buc, E.~B. Fox, and R.~Garnett, eds.), pp.~4083--4092,
  2019.

\bibitem{gnn_np_hard_alpha_go_2019}
K.~Abe, Z.~Xu, I.~Sato, and M.~Sugiyama, ``Solving np-hard problems on graphs
  with extended alphago zero,'' {\em arXiv}, 2019.

\bibitem{learning_comb_opt_graphs_2017}
E.~B. Khalil, H.~Dai, Y.~Zhang, B.~Dilkina, and L.~Song, ``Learning
  combinatorial optimization algorithms over graphs,'' in {\em Advances in
  Neural Information Processing Systems 30} (I.~Guyon, U.~von Luxburg,
  S.~Bengio, H.~M. Wallach, R.~Fergus, S.~V.~N. Vishwanathan, and R.~Garnett,
  eds.), pp.~6348--6358, 2017.

\bibitem{graph_partitioning_ml_2019}
A.~Nazi, W.~Hang, A.~Goldie, S.~Ravi, and A.~Mirhoseini, ``Gap: Generalizable
  approximate graph partitioning framework,'' {\em arXiv}, 2019.

\bibitem{gnn_maximum_independent_set_2018}
Z.~Li, Q.~Chen, and V.~Koltun, ``Combinatorial optimization with graph
  convolutional networks and guided tree search,'' in {\em Advances in Neural
  Information Processing Systems 31} (S.~Bengio, H.~M. Wallach, H.~Larochelle,
  K.~Grauman, N.~Cesa{-}Bianchi, and R.~Garnett, eds.), pp.~537--546, 2018.

\bibitem{gnn_maximum_common_subgraph_2021}
Y.~Bai, D.~Xu, Y.~Sun, and W.~Wang, ``{GLS}earch: Maximum common subgraph
  detection via learning to search,'' in {\em Proceedings of the 38th
  International Conference on Machine Learning, {ICML} 2021, 18-24 July 2021,
  Virtual Event} (M.~Meila and T.~Zhang, eds.), vol.~139 of {\em Proceedings of
  Machine Learning Research}, pp.~588--598, {PMLR}, 2021.

\bibitem{gnn-beam_search_2019}
C.~K. Joshi, T.~Laurent, and X.~Bresson, ``An efficient graph convolutional
  network technique for the travelling salesman problem,'' {\em arXiv}, 2019.

\bibitem{learning_tsp_2022}
C.~K. Joshi, Q.~Cappart, L.-M. Rousseau, and T.~Laurent, ``Learning tsp
  requires rethinking generalization,'' in {\em 27th International Conference
  on Principles and Practice of Constraint Programming (CP 2021)}, Schloss
  Dagstuhl-Leibniz-Zentrum f{\"u}r Informatik, 2021.

\bibitem{prates_2018}
M.~O.~R. Prates, P.~H.~C. Avelar, H.~Lemos, L.~C. Lamb, and M.~Y. Vardi,
  ``Learning to solve np-complete problems: {A} graph neural network for
  decision {TSP},'' in {\em The Thirty-Third {AAAI} Conference on Artificial
  Intelligence, {AAAI} 2019}, pp.~4731--4738, {AAAI} Press, 2019.

\bibitem{pointer_networks_2015}
O.~Vinyals, M.~Fortunato, and N.~Jaitly, ``Pointer networks,'' {\em Advances in
  neural information processing systems}, vol.~28, 2015.

\bibitem{learning_improvement_heuristics_tsp_2019}
Y.~Wu, W.~Song, Z.~Cao, J.~Zhang, and A.~Lim, ``Learning improvement heuristics
  for solving routing problems,'' {\em IEEE transactions on neural networks and
  learning systems}, vol.~33, no.~9, pp.~5057--5069, 2021.

\bibitem{attention_learn_tsp_2018}
W.~Kool, H.~van Hoof, and M.~Welling, ``Attention, learn to solve routing
  problems!,'' in {\em 7th International Conference on Learning
  Representations, {ICLR} 2019, New Orleans, LA, USA, May 6-9, 2019},
  OpenReview.net, 2019.

\bibitem{bengio_2016}
I.~Bello, H.~Pham, Q.~V. Le, M.~Norouzi, and S.~Bengio, ``Neural combinatorial
  optimization with reinforcement learning,'' in {\em 5th International
  Conference on Learning Representations, {ICLR} 2017, Toulon, France, April
  24-26, 2017, Workshop Track Proceedings}, OpenReview.net, 2017.

\bibitem{graph_learning_comb_opt_survey_2020}
Y.~Peng, B.~Choi, and J.~Xu, ``Graph learning for combinatorial optimization: A
  survey of state-of-the-art,'' {\em Data Science and Engineering}, vol.~6,
  pp.~119--141, Jun 2021.

\bibitem{gnn_nsc_survey_2020}
L.~C. Lamb, A.~S. d'Avila Garcez, M.~Gori, M.~O.~R. Prates, P.~H.~C. Avelar,
  and M.~Y. Vardi, ``Graph neural networks meet neural-symbolic computing: {A}
  survey and perspective,'' in {\em Proceedings of the Twenty-Ninth
  International Joint Conference on Artificial Intelligence, {IJCAI} 2020}
  (C.~Bessiere, ed.), pp.~4877--4884, ijcai.org, 2020.

\bibitem{weisfeiler_lehman_graph_kernels}
N.~Shervashidze, P.~Schweitzer, E.~J. van Leeuwen, K.~Mehlhorn, and K.~M.
  Borgwardt, ``Weisfeiler-lehman graph kernels,'' {\em Journal of Machine
  Learning Research}, vol.~12, no.~77, pp.~2539--2561, 2011.

\bibitem{pycsp3}
C.~Lecoutre and N.~Szczepanski, ``{PYCSP3:} modeling combinatorial constrained
  problems in python,'' {\em CoRR}, vol.~abs/2009.00326, 2020.

\bibitem{roth_2007}
A.~E. Roth, T.~Sönmez, and M.~U. Ünver, ``Efficient kidney exchange:
  Coincidence of wants in markets with compatibility-based preferences,'' {\em
  American Economic Review}, vol.~97, pp.~828--851, June 2007.

\bibitem{kep_ip_models_2013}
M.~Constantino, X.~Klimentova, A.~Viana, and A.~Rais, ``New insights on
  integer-programming models for the kidney exchange problem,'' {\em European
  Journal of Operational Research}, vol.~231, no.~1, pp.~57--68, 2013.

\bibitem{pna_2020}
G.~Corso, L.~Cavalleri, D.~Beaini, P.~Liò, and P.~Veličković, ``Principal
  neighbourhood aggregation for graph nets,'' 2020.

\bibitem{gat_v2_2021}
S.~Brody, U.~Alon, and E.~Yahav, ``How attentive are graph attention
  networks?,'' 2021.

\bibitem{saidman_2006}
S.~L. Saidman, A.~E. Roth, T.~S{\"o}nmez, M.~U. {\"U}nver, and F.~L. Delmonico,
  ``Increasing the opportunity of live kidney donation by matching for two- and
  three-way exchanges,'' {\em Transplantation}, vol.~81, no.~5, 2006.

\bibitem{graph_vae}
M.~Simonovsky and N.~Komodakis, ``Graphvae: Towards generation of small graphs
  using variational autoencoders,'' in {\em Artificial Neural Networks and
  Machine Learning - {ICANN} 2018}, vol.~11139 of {\em Lecture Notes in
  Computer Science}, pp.~412--422, Springer, 2018.

\bibitem{molgan}
N.~De~Cao and T.~Kipf, ``Molgan: An implicit generative model for small
  molecular graphs,'' {\em arXiv preprint arXiv:1805.11973}, 2018.

\end{thebibliography}

\end{document}